%% file: iclr2026_conference.tex
\documentclass{article} 
\usepackage{iclr2026_conference,times}
\usepackage{multirow}
\usepackage{xcolor}
\usepackage{colortbl}
\usepackage{graphicx}
\input{math_commands.tex}

\usepackage{hyperref}
\usepackage{url}
\usepackage{makecell}
\usepackage{booktabs}
\usepackage{threeparttable}
\usepackage[utf8]{inputenc}
\usepackage{textcomp}
\usepackage{newunicodechar}
\newunicodechar{μ}{\textmu}

\title{Cannistraci-Hebb Training on Ultra-Sparse Spiking Neural Networks}

\iclrfinalcopy

\author{
\parbox{\textwidth}{\centering
Yuan Hua\textsuperscript{1}, 
Jilin Zhang\textsuperscript{1}, 
Yingtao Zhang\textsuperscript{2, 3, 4}, 
Wenqi Gu \textsuperscript{2, 3, 4}, 
Leyi You\textsuperscript{1}, 
Baobo Xiong\textsuperscript{1}, \\
Carlo Vittorio Cannistraci\textsuperscript{2, 3, 4, 5, *}
and Hong Chen\textsuperscript{1, *}
} \\
\\
\parbox{\textwidth}{\centering
\textsuperscript{1}School of Integrated Circuits, Tsinghua University, Beijing China \\
\textsuperscript{2}Center for Complex Network Intelligence (CCNI) \\
\textsuperscript{3}Tsinghua Laboratory of Brain and Intelligence (THBI) \\
\textsuperscript{4}Department of Computer Science\\ 
\textsuperscript{5}Department of Biomedical Engineering, Tsinghua University, Beijing, China \\
\textsuperscript{*}Corresponding authors \\
}
\\
\parbox{\textwidth}{\centering
\texttt{hongchen@tsinghua.edu.cn}, 
\texttt{kalokagathos.agon@gmail.com}
}
}

\begin{document}

\maketitle
\begin{abstract}
Inspired by the brain's spike-based computation, spiking neural networks (SNNs) inherently possess temporal activation sparsity. However, when it comes to the sparse training of SNNs in the structural connection domain, existing methods fail to achieve ultra-sparse network structures without significant performance loss, thereby hindering progress in energy-efficient neuromorphic computing. This limitation presents a critical challenge: how to achieve high levels of structural connection sparsity while maintaining performance comparable to fully connected networks. To address this challenge, we propose the \underline{C}annistraci-\underline{H}ebb \underline{S}piking \underline{N}eural \underline{N}etwork (CH-SNN), a novel and generalizable dynamic sparse training framework for SNNs consisting of four stages. First, we propose a sparse spike correlated topological initialization (SSCTI) method to initialize a sparse network based on node correlations. Second, temporal activation sparsity and structural connection sparsity are integrated via a proposed sparse spike weight initialization (SSWI) method. Third, a hybrid link removal score (LRS) is applied to prune redundant weights and inactive neurons, improving information flow. Finally, the CH3-L3 network automaton framework inspired by Cannistraci-Hebb learning theory is incorporated to perform link prediction for potential synaptic regrowth. These mechanisms enable CH-SNN to achieve sparsification across all linear layers. We have conducted extensive experiments on six datasets including CIFAR-10 and CIFAR-100, evaluating various network architectures such as spiking convolutional neural networks and Spikformer. The proposed method achieves a maximum sparsity of 97.75\% and outperforms the fully connected (FC) network by 0.16\% in accuracy. Furthermore, we apply CH-SNN within an SNN training algorithm deployed on an edge neuromorphic processor. The experimental results demonstrate that, compared to the FC baseline without CH-SNN, the sparse CH-SNN architecture achieves up to 98.84\% sparsity, an accuracy improvement of 2.27\%, and a 97.5$\times$ reduction in synaptic operations, and the energy consumption is reduced by an average of 55$\times$ across four datasets. To comply with double-blind review requirements, our code will be made publicly available upon acceptance.
\end{abstract}

\section{Introduction}
The increasing computational demands and energy consumption of deep neural networks have spurred the exploration of energy-efficient alternatives. Inspired by the event-driven processing mechanism of the human brain, spiking neural networks (SNNs) have emerged as a promising solution due to their inherent temporal activation sparsity~\citep{snn1,snn2}. The temporal activation sparsity of SNNs stems from their spiking characteristics—neurons only fire spikes when the membrane potential reaches the threshold, remaining in a resting state for most of the time~\citep{WU2023126247}. Compared to artificial neural networks (ANNs), SNNs demonstrate significant advantages in energy efficiency, making them well-suited for a variety of edge-side applications such as gas detection~\citep{anp-g}, sEMG-based gesture recognition~\citep{anp-i} and real-time multi-object recognition~\citep{multi}.

Despite the inherent advantages in temporal activation sparsity, SNNs often suffer from a fixed architecture that lacks flexibility and structural plasticity, which limits the learning capability of SNNs and their application in resource-limited neuromorphic hardware~\citep{loihi}. To address this problem, previous works introduce structural connection sparsity through network pruning and regrowth~\citep{prune}. Although sparse training has proven effective in reducing parameter counts and improving computational efficiency in ANNs~\citep{dst1, Rigl, dst3, dst4}, its application in SNNs remains challenging. This is due to the spike-based computation and the non-differentiable nature of the spiking activation function of SNNs, which hinder direct gradient-based optimization. Consequently, most sparse training methods for ANNs that rely on gradient information cannot be directly applied to SNNs.

Specifically, current research in sparse SNNs training methods faces a significant challenge in achieving high levels of structural connection sparsity while maintaining performance comparable to that of their fully connected counterpart. For instance, the adaptive structural development of SNN (SD-SNN) model~\citep{HAN2025121481}, incorporates multiple brain-inspired developmental mechanisms, including synaptic elimination, neuronal pruning and synaptic regeneration. Besides, it also uses adaptive pruning and regrowth rates, which led to the structural stability. As a result, SD-SNN achieves 98.56\% accuracy with 1.45\% improvement on DVS-Gesture dataset, but only reaches a maximum sparsity of 61.10\%. Similarly, ~\cite{shen2025improvingsparsestructurelearning} propose a two-stage dynamic structure learning method, effectively addressing the limitations of fixed pruning ratios and static sparse training methods prevalent in existing models. Nonetheless, their approach attains an average structural connection sparsity of around 70\%. Some studies directly apply ANN-based sparse training methods to SNNs, Gradient Rewiring~\citep{shen2025improvingsparsestructurelearning} method achieves up to 90\% sparsity. However, it exhibits an accuracy degradation of 3.55\% compared to its fully connected counterpart.

To address this challenge, we propose the \underline{C}annistraci-\underline{H}ebb \underline{S}piking \underline{N}eural \underline{N}etwork (CH-SNN), a novel and generalizable dynamic sparse training framework for SNNs, which achieves high levels of structural connection sparsity and maintaining performance comparable to that of its fully connected (FC) counterpart. The main contributions of this work are summarized as follows:

\begin{itemize}
    \item \textbf{Introducing a novel sparse training framework.} We propose a four-stage dynamic sparse training framework (CH-SNN) consisting of sparse topology initialization, sparse weight initialization, network pruning and network regrowth. CH-SNN attains 99\% structural connection sparsity in all linear layers and shows better performance than FC networks on the CIFAR-100, MNIST, N-MNIST, CIFAR10-DVS and DVS-Gesture datasets respectively.
    \item \textbf{Proposing efficient initialization methods.} We propose two initialization methods, one is Sparse Spike Correlated Topological Initialization (SSCTI) which initializes an ultra-sparse network structure by leveraging correlations among input nodes, another is Sparse Spike Weight Initialization (SSWI) which incorporates temporal activation sparsity and structural connection sparsity of SNNs to initialize weights. SSCTI and SSWI enhance the performance of the link predictor and facilitate faster training from the initial phases.
    \item \textbf{Demonstrating superior performance across architectures and datasets.} We have conducted extensive experiments, the experimental results demonstrate that CH-SNN outperforms existing sparse SNN training methods across six datasets (CIFAR-10, CIFAR-100, MNIST, N-MNIST, CIFAR10-DVS and DVS-Gesture) and three network structures. Notably, it attains a 0.16\% accuracy improvement over the FC network at a sparsity of 97.75\%. We apply CH-SNN to a hardware-friendly algorithm S-TP, which has been implemented on a neuromorphic processor for edge-side AI applications. Experimental results show that CH-SNN significantly improve energy efficiency, achieving an average improvement of 55$\times$ across four datasets.
\end{itemize}

\section{Related works}

\subsection{Sparse spiking neural networks}
Structural connection sparsification is one of the key technologies for enhancing SNNs energy efficiency. By reducing redundant links and neurons within the model, it can significantly reduce computational and storage overhead. Existing sparse SNNs training methods can be categorized into pruning and sparse training.

\textbf{Pruning.} Pruning methods initialize a fully connected network structure and gradually remove insignificant links during training. Current SNNs pruning approaches can be divided into the two types: (1) Biological plasticity pruning. This approach draws inspiration from the developmental mechanisms of the brain, leveraging biological synaptic plasticity to accomplish the pruning of SNNs. \cite{hanbing_tpami, HAN2025121481} propose the developmental plasticity-inspired adaptive pruning method, which takes into account multiple biologically realistic mechanisms, so that the network structure can be dynamically optimized. \cite{rw_stdp} present a sparse SNN training method where pruning is based on the spike timing dependent plasticity model (STDP). Links between pre-neuron and post-neuron with low correlation or uncorrelated spiking activity are pruned. \cite{DYNSNN} propose a dynamic pruning framework (named DynSNN) for SNNs, enabling dynamic optimization of the network topology. (2) Transfer ANNs pruning method to SNNs. These methods adapt pruning techniques from ANNs. For instance, \cite{rw_yanqi_2022} use different functions describing the growing threshold of state transition to regulate the pruning speed, avoiding disastrous performance degradation at the final stage of training. \cite{admm} formulate the link pruning problem as a constrained optimization problem, which is addressed by integrating spatiotemporal backpropagation (STBP) with the alternating direction method of multipliers (ADMMs). Backpropagation with sparsity regularization (BPSR)~\citep{rw_bpsr} incorporates an $L_1$ regularization term into the loss function to drive the weights toward zero, followed by a static threshold-based pruning method, thereby achieving network structural connection sparsification.

\textbf{Sparse training.} In contrast to pruning methods, sparse training begins with a sparsely connected network and dynamically alternates between pruning less important connections and growing new ones during learning. It maintains sparsity in both the forward and backward propagation during the training process, resulting in lower hardware requirements. For example, the Deep Rewiring (Deep R)~\citep{rw_bellec2018deeprewiringtrainingsparse} method prunes links when their value changes sign during updates, and randomly regenerate an equivalent number of links. This process is repeated over multiple rounds. Based on Deep R, \cite{gradR} introduce a Gradient Rewiring (Grad R) approach to further modify the gradient values of the links, enabling previously pruned links to regenerate. Furthermore, ~\cite{shen2025improvingsparsestructurelearning} propose a two-stage dynamic structure learning method for deep SNNs, the first stage evaluates the network's compressibility based on the PQ index~\citep{pq_index} and adaptively determines the regrowth ratio, and the second stage performs pruning and regrowth according to this ratio. \cite{rw_snn-cg} propose a spiking neural network with connection gates (SNN-CG) to jointly learn the topology and the weights in SNN. The connection structures and the weights are learned alternately until a termination condition is satisfied. Neurogenesis dynamics-inspired spiking neural network (NDSNN) training method~\citep{rw_dac_dst} trains a model from scratch using dynamic sparsity. NDSNN creates a drop-and-grow strategy to promote link reduction. Based on RigL~\citep{Rigl},~\cite{Srigl} propose a sparse-to-sparse dynamic sparse training method named Structured RigL (SRigL). SRigL learns a SNN with constant fan-in fine-grained structured sparsity while maintaining generalization comparable with RigL. 

\subsection{Cannistraci-Hebb theory and network topology intelligence}
Inspired by the dynamic sparse connectivity characteristics of the brain, the Cannistraci-Hebb (CH) theory~\citep{carlo1, carlo2, carlo3, carlo4, neurips2025} is a general theoretical framework developed in the field of network science to predict the non-observed dynamic connectivity of complex networks, using the mere knowledge of the network topology.  CH theory is also recently introduced~\citep{neurips2025, iclr2024} for dynamic sparse training for deep AI, demonstrating a gradient-free link regrowth mechanism that relies solely on topological information. For example, Cannistraci-Hebb Training (CHT)~\citep{cht} is applied to ANNs, utilizing the CH3-L3 network automaton for link prediction. CH3-L3 is one of the highest-performing and most robust network automata under the Cannistraci-Hebb theory~\citep{neurips2025}. It can automatically evolve the network topology of a given structure by identifying node pairs with the fewest external connections within the local community structure, thereby guiding link regrowth. For multiple tasks, CHT achieves better performance surpassing than fully connected networks with only 1\% connections, demonstrating the ultra-sparse advantage. Importantly, CHT is shown to induce during training also a node sparsification process (called network node percolation), which at the end of the training compressed the node size of certain networks to around the 30 percent of the initial size, preserving or improving task performance. Furthermore,~\cite{chts} put forward a Cannistraci-Hebb Training soft rule (CHTs) which probabilistically removes network links based on a removal fraction and regrows new links according to CH3-L3 prediction scores, overcoming CHT's tendency to fall into epitopological local minima during the early stages of training when topological noise is significant.

\section{Methods}
\subsection{Spiking neural network}
\textbf{Fundamentals.} Unlike artificial neural networks, SNNs use sparse spike signals to transmit information. The spike signals enable SNNs to avoid Multiply-Accumulate (MAC) operations. Thereby reducing energy consumption and computational load~\citep{METHOD_SNN.2017.00682}. In this paper, we adopt the leaky integrate-and-fire (LIF) neuron~\citep{lif} to process spike signals, as shown in Equation~(\ref{equ: lif}).
\begin{equation}
\label{equ: lif}
v_j(t+1) = (1-z_j(t))\alpha v_j(t) + \sum_i \emW_{ij}\evx_i(t+1), \quad z_j(t) = U(v_j(t)-\theta)
\end{equation}
where $t$ denotes the time step, $v_j(\cdot)$ represents the membrane potential of the neuron $j$, $\alpha$ is the membrane potential decay constant, $\emW_{ij}$ denotes the synaptic weight, $\evx_i(t)$ is the input spike, and $U$ is the step function. When the membrane potential accumulates and exceeds the firing threshold $\theta$, the neuron emits an output spike, denoted by $z_j(\cdot)=1$. Otherwise, the neuron remains silent, i.e., $z_j(\cdot)=0$. After firing a spike, the membrane potential is reset to zero.

\textbf{Training method.} It is challenging to apply standard gradient descent to SNNs. The step function in Equation~(\ref{equ: lif}) results in a derivative that is zero almost everywhere and undefined at the threshold point. This makes the direct calculation of partial derivatives $\frac{\partial z_j(\cdot)}{\partial v_j(\cdot)}$ impossible using conventional calculus, which prevents the application of the backpropagation algorithm. Thus we use the surrogate gradient method~\citep{stbp} to update the weights of SNNs. 

\textbf{Sparse Target Propagation.} Sparse Target Propagation (S-TP)~\citep{anp-i} adopts a hardware-friendly surrogate gradient method. S-TP randomly selects target windows in the learning process, reducing over 90\% of the spike number in the learning process without noticeable accuracy degradation. S-TP has been implemented in a low-power neuromorphic processor, which proves its notable hardware-friendliness.

\subsection{Cannistraci-Hebb Spiking Neuron Network}
\begin{figure}
    \centering
    \includegraphics[width=1\linewidth]{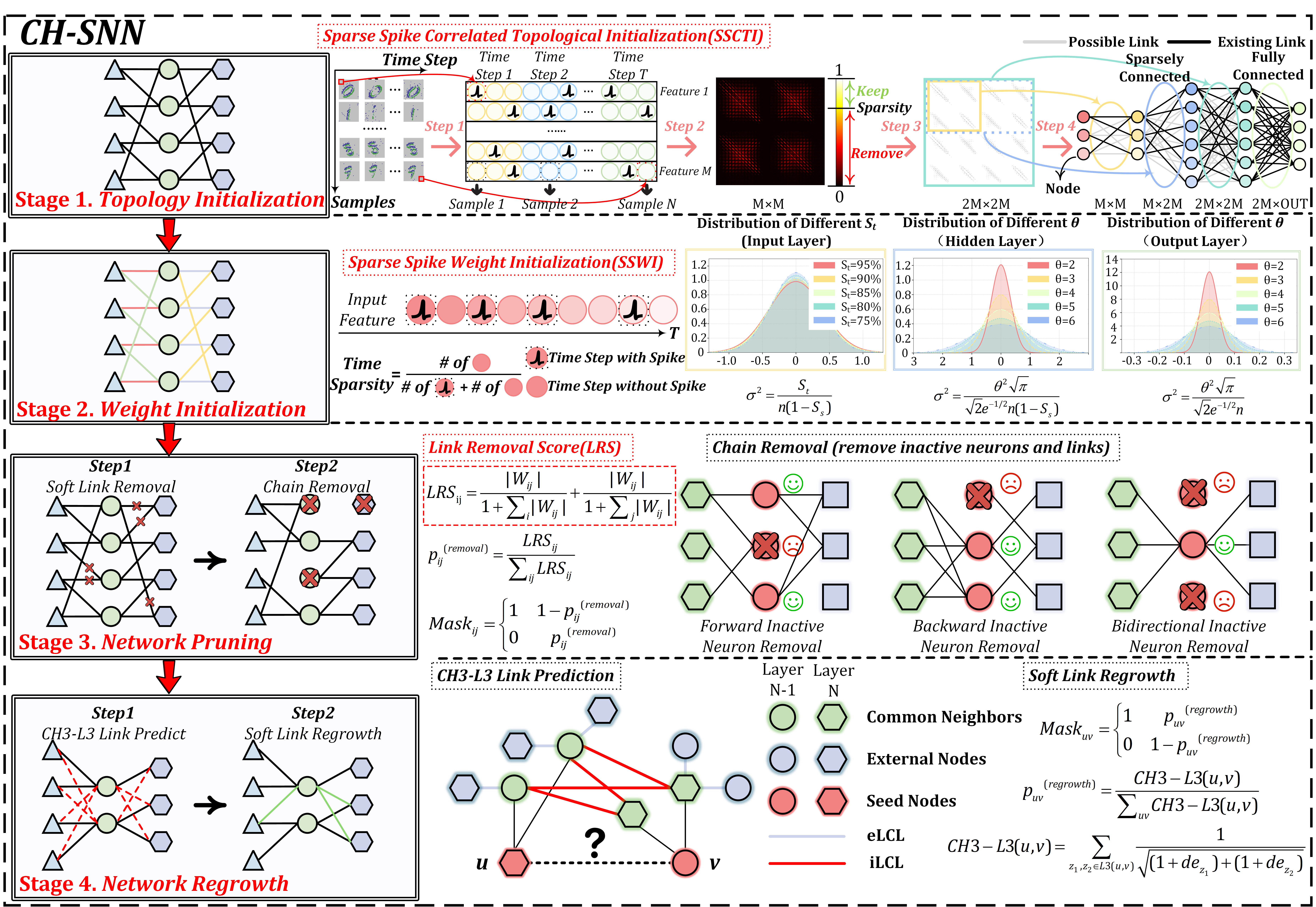}
    \caption{The framework of Cannistraci-Hebb Spiking Neural Network (CH-SNN).}
    \label{fig:CSTI}
\end{figure}

In this paper, we propose the four-stage dynamic sparse training method for SNNs, named Cannistraci-Hebb Spike Neuron Network (CH-SNN). It is a general sparse training framework capable of sparsifying all linear layers in SNNs. The framework of CH-SNN is illustrated in Figure~\ref{fig:CSTI}. \textbf{The first stage is sparse topology initialization.} We propose a sparse topology initialization method named Sparse Spike Correlated Topological Initialization (SSCTI) based on Pearson’s phi coefficient to initialize an ultra-sparse network. \textbf{The second stage is sparse weight initialization.} We introduce the Sparse Spike Weight Initialization (SSWI) method, which incorporates the temporal activation sparsity, structural connection sparsity and neuronal threshold information of SNNs into the weight initialization process to perform weight initialization. \textbf{The third stage is network pruning.} We use a probability-based links pruning strategy to remove links according to a dynamic ratio $\zeta$. Subsequently, inactive neurons are identified and removed, according to the CHT network percolation procedure ~\citep{cht}. \textbf{The fourth stage is network regrowth.} Here, the regrowth score of potential links is computed using CH3-L3. According to this score, links are regenerated with the same ratio applied in the network pruning stage, thereby maintaining the predefined structural connection sparsity. Links with higher regrowth scores are proportionally sampled according to the CHTs methodology ~\citep{chts}. 

\subsubsection{Sparse topology initialization}
As we know, the topology of a network should reflect the relationships between node features within some latent geometric space~\citep{cannistraci2020geometricalcongruenceefficientgreedy}. The correlations between input features directly define the geometric relationships among nodes in this latent feature space. Therefore, by computing the correlations between nodes in the input layer, we preserve connections between highly correlated nodes according to the predefined sparsity. Thus, we propose the Sparse Spike Correlated Topological Initialization (SSCTI) method, as shown in Figure~\ref{fig:CSTI} Stage 1. Since the input of SNNs are discrete binary spike trains, we measure the correlation between input nodes using Pearson’s phi coefficient~\citep{pearson2015theory}, which measures the strength and direction of association between two binary variables. We take each dimension of the input data $x_i$ and each time step $t$ as a variable and an independent sample, respectively. Thus, the total number of samples is $N\times T$. The Pearson’s phi coefficient is described in Equation~(\ref{equ:phi}). 
\begin{equation}
\label{equ:phi}
    \phi_{ij} = \sqrt{\frac{\chi^2_{ij}}{2NT}} = \sqrt{\frac{\sum_{t=1}^{NT}(x_i(t)-E_i)^2/E_i + \sum_{t=1}^{NT}(x_j(t)-E_j)^2/E_j}{2NT}}
\end{equation}
where $\phi_{ij}$ represents the Pearson's phi coefficient between input data $x_i$ and $x_j$, $M$ denotes the dimension of the input data, $T$ is total time step, $N$ stands for the number of samples, $\chi_{ij}^2$ represents the Chi-square statistic, and $E_i$ is the mean value of $x_i$. From Equation~(\ref{equ:phi}), we obtain the correlation matrix $\Phi \in \mathbb{R}^{M\times M}$, we keep the top ($1-S_s$) proportion of links of the SNN with the strongest correlations and remove other links in SNN, where $S_s$ stands for the structural connection sparsity, thereby completing the initialization of the network structure. The dimensionality of the hidden layer is determined by an expansion factor $\beta \geq 1,\  \beta \in \mathbb{Z}$, such that the dimension of the hidden layer equals the input dimension multiplied by $\beta$. This allows the hidden layer dimensionality to be flexibly adjusted, as shown in Figure~\ref{fig:CSTI} Stage 1. However, when CH-SNN is used to sparsify intermediate layers, such as linear layers within spiking convolutional neural networks~\citep{scnn} or Spikformer~\citep{spikformer}, the input distribution may be altered by preceding convolutional layers or attention layers. This makes it difficult for SSCTI to accurately capture feature correlations between nodes. To address this issue, we adopt a uniform random initialization strategy, which ensures that each node retains an equal number of connections. 
\subsubsection{Sparse spike weight initialization}
 For ANNs, weight initialization strategies such as Kaiming initialization~\citep{kaiming} are widely adopted. Most of these methods assume that weights follow a zero-mean Gaussian distribution, and determine the variance of this distribution under the principle of maintaining consistent variance of input data across layers. However, such approaches cannot be directly applied to the weight initialization in structural sparse networks. Although methods like SWI~\citep{cht} have been proposed for sparse artificial neural networks, they are unsuitable for SNNs owing to their inability to incorporate temporal activation sparsity and the unique activation function of LIF neurons. To address this problem, we put forward the Sparse Spike Weight Initialization (SSWI) method, which incorporates the temporal activation sparsity ($S_t$), structural connection sparsity ($S_s$) and the neuronal threshold information ($\theta$) of SNNs into the weight initialization process. The detailed derivation is provided in Appendix ~\ref{a.1}. The SSWI method is presented in Equation~(\ref{equ:sswi}).
\begin{equation}
\label{equ:sswi}
    SSWI(\emW_{ij}^{(l)}) \sim \mathcal{N}(0,\sigma^2),\quad \sigma^2 = \left\{
\begin{aligned}
&\frac{S_t}{n(1-S_s)}, \quad (l=1) \\
&\frac{\theta^2\sqrt{\pi}}{\sqrt{2}e^{-1/2}n(1-S_s)},\quad ( 1<l<L) \\
& \frac{\theta^2\sqrt{\pi}}{\sqrt{2}e^{-1/2}n},\quad (l=L)
\end{aligned}
\right.
\end{equation}
where $S_t$ denotes the temporal activation sparsity of the input data in SNNs, $S_s$ represents the structural connection sparsity, $l$ is the index of the layer (with a total of $L$ layers), $n$ indicates the input feature dimension of the $l$ layer, and $\theta$ denotes the spike threshold in the LIF neuron. SSWI enhances training efficiency, leading to faster convergence from the initial phases.

\subsubsection{Network pruning}
\textbf{Link Removal.} We propose a hybrid strategy to calculate the link removal score ($LRS$) that combines relative importance (RI) and weight magnitude (WM). The approach not only accelerates network sparsification but also promotes the activation of more neurons during training. The $LRS$ is defined as follows:

\begin{equation}
\label{equ:ri}
    LRS_{ij}^{(l)}=\frac{|W_{ij}^{(l)}|}{1+\sum_i |W_{ij}^{(l)}|} + \frac{|W_{ij}^{(l)}|}{1+\sum_j |W_{ij}^{(l)}|}
\end{equation}
where $LRS_{ij}^{(l)}$ denotes the link removal score of the weight $W_{ij}^{(l)}$, $\sum_j |W_{ij}^{(l)}|$ represents the sum of the magnitude of all weights connected to the input neuron $i$, and $\sum_i |W_{ij}^{(l)}|$ denotes the sum of the magnitude of all weights connected to the output neuron $j$. Instead of using the magnitude of the $LRS$ as the direct criterion for link removal, we sample from a multi-nomial distribution based on the $LRS$ value to determine whether a link should be removed.

\textbf{Chain Removal.} 
After link removal, neurons that are unilaterally or bilaterally disconnected (i.e., without any incoming or outgoing links) are regarded as inactive neurons. Since these inactive neurons lose the ability to transmit information, they may hinder information flow throughout the network. Because of the mechanism of CH3-L3, such inactive neurons are unable to regrow new links during the network regrowth stage. Therefore, during the chain removal step, we permanently remove them from the network. As illustrated in Figure~\ref{fig:CSTI} Stage 3, this process enhances overall information propagation.

\subsubsection{Network regrowth}
We employ CH3-L3 to compute the link regrowth score for potential links, as CH3-L3 is recognized as the most robust and stable link predictor within the Cannistraci-Hebb theory~\citep{iclr2024, neurips2025, Brain_network_science}. To mitigate the risk of falling into epitopological local minima due to structural noise in the network, we sample from a binomial distribution based on the regrowth score to stochastically determine whether a link should be regenerated, instead of using the regrowth score directly as the criterion for link regrowth~\citep{chts}. The formula for calculating the link regrowth score is as follows:

\begin{equation}
\label{equ:ri}
    \textbf{CH3-L3}(u, v)= \sum_{z_1,z_2\in l3(u,v)}\frac{1}{\sqrt{(1+de_{z_{1}})\times (1+de_{z_{2}})}}
\end{equation}

where $u$ and $v$ are two nodes that may potentially form a link, and $z_1$, $z_2$ denote two intermediate nodes along a path of length 3 between $u$ and $v$—also referred to as common neighbor nodes of $u$ and $v$. The terms $de_{z_1}$ and $de_{z_2}$ represent the external local community connectivity degrees of nodes $z_1$ and $z_2$, respectively. A detailed description of the CH3-L3 is provided in Appendix ~\ref{a.2}.

\section{Experiments}

\subsection{Comprehensive performance comparison with other methods}
\begin{figure}[h]
    \centering
    \includegraphics[width=1\linewidth]{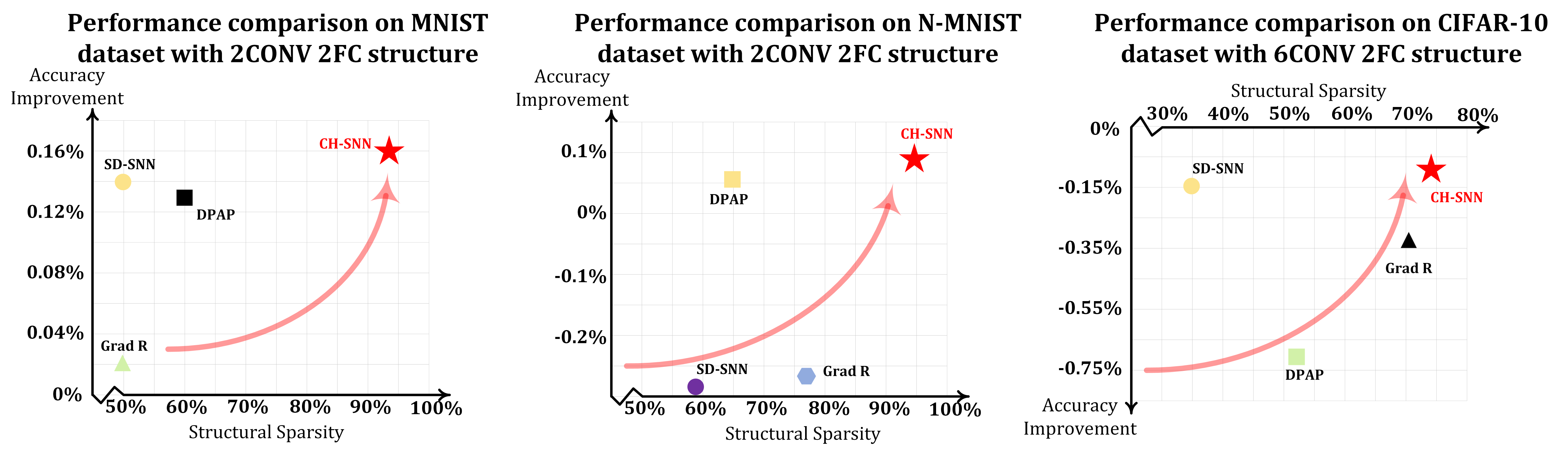}
    \caption{Performance comparison of different methods on MNIST, N-MNIST and CIFAR-10. We plot the performance of different sparse SNN training methods with structural sparsity on the x-axis and accuracy improvement on the y-axis. The plot clearly shows that CH-SNN achieves the highest level of sparsity alongside the greatest improvement in accuracy.}
    \label{fig:acc-sparsity}
\end{figure}

We compare our CH-SNN with existing sparse SNNs training methods including Grad R~\citep{gradR}, SD-SNN~\citep{HAN2025121481} and DPAP~\citep{hanbing_tpami}, using the same network architectures for fair comparison. In addition, we have conducted experiments on the Spikformer~\citep{spikformer} architecture to further verify our methods. It is worth noting that none of the compared methods have been evaluated on the Spikformer. Detailed experimental settings are provided in Appendix~\ref{a.setup}. The experimental results are summarized in Table~\ref{tab:performance} and Figure~\ref{fig:acc-sparsity}.
\begin{table}[h] 
\caption{Performance comparison of different methods on  CIFAR-10, CIFAR-100, MNIST, N-MNIST, CIFAR10-DVS and DVS-Gesture  datasets. The gray section indicates the performance of CH-SNN. For each dataset, we have bolded the method with the highest accuracy improvement and the one with the highest sparsity. We exclude Spikformer from our comparison here.}
\centering
\renewcommand{\arraystretch}{1.0}
\label{tab:performance}
\begin{tabular}{lllccc}
\toprule[1.5pt]
\textbf{Dataset} & \textbf{Method} & \textbf{Network} & \textbf{Link sparsity}  & \textbf{Acc.} & \makecell{\textbf{Accuracy} \\ \textbf{improvement}} \\
\midrule[0.8pt] 
\multirow{5}{*}{\textbf{CIFAR-10}} 
                    & Grad R            & 6Conv 2FC             & 71.59\%               & 92.54\%               & --0.30\% \\
                    & DPAP              & 6Conv 2FC             & 50.80\%               & 93.83\%               & --0.71\% \\
                    & SD-SNN            & 6Conv 2FC             & 35.57\%               & 94.59\%               & --0.15\% \\
\rowcolor{gray!15}  & \textbf{CH-SNN}   &\textbf{6Conv 2FC}     & \textbf{74.62\%}  &\textbf{94.60\%}       &\textbf{--0.14\%} \\
\rowcolor{gray!15}  & CH-SNN   & Spikformer   &82.21\%   & 94.26\%      &--0.10\%   \\
\hline

\multirow{3}{*}{\textbf{CIFAR-100}} 
                   & \textbf{SD-SNN}            & \textbf{6Conv 2FC}              & \textbf{36.94\%}               & \textbf{75.33\%}               & \textbf{+3.27\%} \\
\rowcolor{gray!15}  \textbf{CIFAR-100} & \textbf{CH-SNN}   & \textbf{6Conv 2FC}     & \textbf{74.45\%}  & \textbf{75.22\%}      & \textbf{+3.16\%}  \\
\rowcolor{gray!15} & CH-SNN   & Spikformer    & 82.11\%  & 76.23\%      & +0.75\%   \\
\hline
\multirow{9}{*}{\textbf{MNIST}} 
                    & Grad R            & 2FC                   & 74.29\%               & 98.59\%               & --0.33\% \\
                    & DPAP              & 2FC                   & 77.36\%               & 98.74\%               &  --0.07\% \\
                    & SD-SNN            & 2FC                   & 45.86\%               & 98.90\%               &  +0.09\% \\
\rowcolor{gray!15}  & \textbf{CH-SNN}   & \textbf{2FC}          &  \textbf{97.75\%} & \textbf{98.97\%}      & \textbf{+0.16\%}   \\
                    & Grad R            & 2Conv 2FC             & 49.16\%               & 99.37\%               & +0.02\% \\
                    & DPAP              & 2Conv 2FC             & 61.25\%               & 99.59\%               & +0.13\% \\
                    & SD-SNN            & 2Conv 2FC             & 49.83\%               & 99.51\%               & +0.14\% \\
\rowcolor{gray!15}  & \textbf{CH-SNN}   & \textbf{2Conv 2FC}    &  \textbf{93.91\%} & \textbf{99.53\%}      &\textbf{+0.16\%} \\
\rowcolor{gray!15}  & CH-SNN   & Spikformer   &  81.72\% & 99.73\%      & +0.02\%   \\
\hline

\multirow{5}{*}{\textbf{N-MNIST}} 
                    & Grad R            & 2Conv 2FC             & 75.00\%               & 98.56\%               & --0.27\% \\
                    & DPAP              & 2Conv 2FC             & 63.95\%               & 99.59\%               & +0.06\% \\
                    & SD-SNN            & 2Conv 2FC             & 58.62\%               & 98.78\%               & --0.29\% \\
\rowcolor{gray!15}  & \textbf{CH-SNN}   & \textbf{2Conv 2FC}    & \textbf{94.73\%}  & \textbf{99.15\%}      & \textbf{+0.08\%} \\
\rowcolor{gray!15}  & CH-SNN   & Spikformer   & 85.74\%      & 99.45\%      & +0.10\%   \\
\hline
\rowcolor{gray!15} \textbf{CIFAR10-DVS}  & \textbf{CH-SNN}   & \textbf{6Conv 2FC}    & \textbf{84.34\%}      & \textbf{72.00\%}      & \textbf{+1.50\%} \\
\rowcolor{gray!15}  & CH-SNN   & Spikformer    & 85.37\%     & 70.60\%    & +0.40\% \\
\hline
\multirow{5}{*}{\textbf{DVS-Gesture}} 
                    & Deep R            & 2Conv 2FC             & 75.00\%               & 81.23\%               & --2.89\% \\
                    & \textbf{Grad R}            & \textbf{2Conv 2FC}             & \textbf{75.00\%}               & \textbf{91.95\%}               & \textbf{+7.83\%} \\
                    & SD-SNN            & 2Conv 2FC             & 61.10\%               & 96.21\%               & +1.14\% \\
\rowcolor{gray!15}  & \textbf{CH-SNN}   & \textbf{2Conv 2FC}    & \textbf{94.73\%}  & \textbf{95.45\%}      & \textbf{+0.38\%} \\
\rowcolor{gray!15}  & CH-SNN   & Spikformer   & 82.25\% & 93.56\%     & +1.14\%   \\
\bottomrule[1.5pt] 
\end{tabular}
\end{table}

\textbf{Performance on Non-Spiking Datasets.} On the MNIST dataset with the two-layer fully connected (2FC) architecture, CH-SNN attains a 97.75\% sparsity while improving accuracy by 0.16\% over the FC baseline. Compared to the state-of-the-art method DPAP, CH-SNN not only increases sparsity by approximately 20\% but also achieves a performance gain of 0.23\%. With the 2CONV2FC architecture, CH-SNN realizes 93.91\% sparsity with a 0.16\% improvement in accuracy. These results demonstrate that CH-SNN delivers both the highest level of sparsity and the most significant performance gains among all compared methods. Furthermore, when applied to the Spikformer architecture, CH-SNN achieves an 81.72\% sparsity with a slight performance improvement of 0.02\%. On the CIFAR-10 dataset, CH-SNN again achieves the highest level of sparsity—74.62\%—with only a minimal accuracy drop of 0.14\%. Similarly, when applied to Spikformer, it maintains a sparsity of 82.21\% with a negligible performance degradation of 0.10\%. On the CIFAR-100 dataset, although the performance improvement of CH-SNN is marginally lower than that of SD-SNN, it increases sparsity by nearly 38\%. Additionally, with the Spikformer architecture, CH-SNN provides a 0.75\% performance improvement at 82.11\% sparsity.

\textbf{Performance on Spiking Datasets.} On the N-MNIST dataset, CH-SNN attains a 0.08\% performance improvement at 94.73\% sparsity, outperforming the FC network. On the DVS-Gesture dataset, although CH-SNN's accuracy gain is lower than those of SD-SNN and Grad R, it still demonstrates 95.45\% accuracy with 0.38\% improvement compared to the FC baseline and reaches significantly higher sparsity level (94.73\%) than all other compared methods. On the CIFAR10-DVS dataset, CH-SNN exhibits an accuracy improvement of 1.50\% at 84.34\% sparsity. Since the CH-SNN framework removes inactive neurons, the sparse network with CH-SNN achieves higher node sparsity compared to other methods. Detailed results are provided in Table~\ref{tab:node_sparsity} of the appendix~\ref{a.node_sparsity}.

\subsection{Experiments on hardware-friendly algorithms S-TP}
\begin{figure}[h]
    \centering
    \includegraphics[width=1\linewidth]{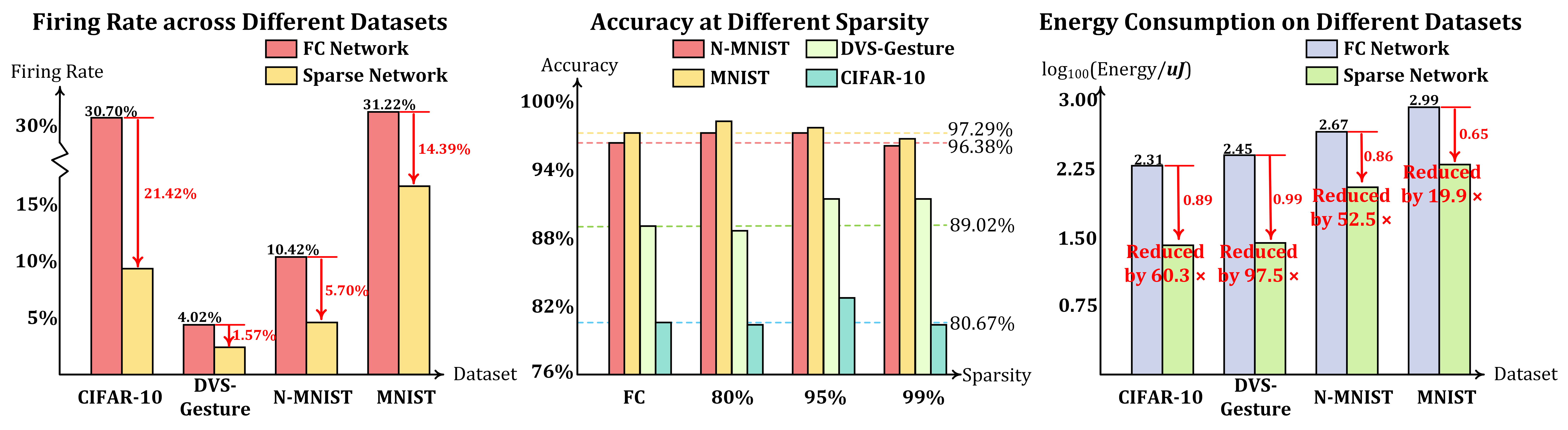}
    \caption{Experimental results of CH-SNN on hardware-friendly algorithm S-TP. A comparison of firing rates between the sparse network with CH-SNN and the FC network across four datasets is presented in the left plot. The middle plot provides an accuracy comparison among the FC network and sparse networks at sparsity levels of 80\%, 95\%, and 99\%. The right plot displays the energy consumption comparison, with the vertical axis on a base-100 logarithmic scale. A scale difference of 0.99 signifies that the energy consumption of the FC network is 97.5 ($100^{0.99}$) times greater than that of the sparse network.}
    \label{fig:energy-scale}
\end{figure}

Hardware-friendly algorithm S-TP has been realized in a chip ANP-I~\citep{anp-i}, which is a low-power neuromorphic processor for edge-side AI applications. To verify our methods, we apply CH-SNN to S-TP. Detailed experimental configurations and network architectures are provided in Appendix~\ref{a.setup}. We have conducted comprehensive experiments to evaluate the performance of CH-SNN in terms of accuracy and energy efficiency. The results are shown in Table~\ref{tab:stp} and Figure~\ref{fig:energy-scale}. 

\textbf{Accuracy Analysis.} At an average sparsity of 96.4\%, the sparse network with CH-SNN achieves the comparable accuracy of the FC network without CH-SNN on all four datasets. Notably, on the DVS-Gesture dataset, sparse network attains a 2.27\% improvement in accuracy at 98.84\% sparsity. For the N-MNIST dataset, although the accuracy of the sparse network is slightly lower than that of the FC network, it successfully prunes nearly half of the nodes (41.90\% node sparsity) with only a minimal accuracy drop of 0.18\%. 

\textbf{Energy Analysis.} We evaluate the energy efficiency in terms of the average firing rate, total spike count, and the number of Synaptic Operations (SOPs). The measured energy consumption of ANP-I chip is 1.5 pJ/SOP~\citep{anp-i}, which is regarded as a baseline. We calculate the total energy consumption by multiplying this baseline by the total SOPs. As illustrated in Figure~\ref{fig:energy-scale}, on the DVS-Gesture dataset, the CH-SNN with sparse connectivity consumes 97.5 times less energy than its fully connected counterpart. Furthermore, it yields an average reduction in energy consumption of 55 times and a 10.77\% decrease in the average firing rate across the four datasets.

\begin{table}[h] 
\caption{Performance and energy consumption of CH-SNN on MNIST, DVS-Gesture, N-MNIST and CIFAR-10 datasets (S-TP). For each dataset, the first row shows the performance of the fully connected network, and the second row shows that of the sparse network with CH-SNN. A 3FC architecture is consistently employed across all experiments, with details provided in the appendix~\ref{a.setup}.}
\centering
\renewcommand{\arraystretch}{0.9}
\label{tab:stp}
\begin{tabular}{llllcccr}
\toprule[1.5pt]
\textbf{Dataset} & \makecell{\textbf{Spike} \\ \textbf{count}} & \textbf{SOPs} & \makecell{\textbf{Firing} \\ \textbf{rate}} & \makecell{\textbf{Link} \\ \textbf{sparsity}} & \makecell{\textbf{Node} \\ \textbf{sparsity}} & \textbf{Acc.} & \textbf{Energy} \\
\midrule[0.8pt] 
\multirow{2}{*}{\textbf{MNIST}} 
                    & $6.1\times 10^{8}$                        & $6.3\times 10^{11}$                & 31.22\%                     & 0\%                   & 0\%                 & 97.29\%             & 948mJ              \\
                    & $\mathbf{3.3\times 10^8}$                         & $\mathbf{3.2\times 10^{10}}$                & \textbf{16.83\%}                     & \textbf{94.59\%}               & \textbf{23.47\%}                 & \textbf{97.56\%}             & \textbf{48mJ}             \\
                    
\midrule[0.8pt] 
\multirow{2}{*}{\makecell{\textbf{DVS-} \\ \textbf{Gesture}}} 
                    & $2.2\times 10^7$        & $5.2\times 10^{10}$                & 4.02\%                & 0\%                   & 0\%               & 89.02\%       & 78mJ             \\
                    & $\mathbf{1.3\times 10^7}$        & $\mathbf{5.0\times 10^8}$                & \textbf{2.45\%}                & \textbf{98.84\%}               & \textbf{12.30\%}           & \textbf{91.29\%}        & \textbf{0.8mJ}          \\
\midrule[0.8pt] 
\multirow{2}{*}{\textbf{N-MNIST}} 
                    & $2.5\times 10^8$        & $1.4\times 10^{11}$                & 10.42\%               & 0\%                   & 0\%               & 96.38\%       & 216mJ           \\
                    & $\mathbf{1.2\times 10^8}$        & $\mathbf{2.9\times 10^9}$                & \textbf{4.72 \%}               & \textbf{98.46\%}               & \textbf{41.90\%}           & \textbf{96.20\%}       & \textbf{4.4mJ}       \\
\midrule[0.8pt] 
\multirow{2}{*}{\textbf{CIFAR-10}} 
                    & $6.3\times 10^7$        & $2.8\times10^{10}$               & 30.70\%               & 0\%                   & 0\%               & 80.67\%       & 41mJ           \\
                    & $\mathbf{1.9\times10^7}$       & $\mathbf{4.8\times10^8}$                & \textbf{9.28\%}              & \textbf{93.78\%}               & \textbf{0\%}           & \textbf{82.84\%}       & \textbf{0.7mJ}       \\
\bottomrule[1.5pt] 
\end{tabular}
\end{table}

\textbf{Supplementary Experiments.} We perform ablation studies to verify the SSCTI and SSWI methods, with results in Appendix~\ref{a.ablation}. Besides, sensitivity analyses are conducted on critical hyperparameters—including learning rate, batch size and pruning ratio—as summarized in Appendix~\ref{a.sensitivity}. A discussion of limitations of the study is provided in the appendix ~\ref{a.limitation}. Details of large language model usage in the writing process can be found in the appendix~\ref{a.llm}.

\section{Conclusion}
 To address the challenge in achieving high levels of structural connection sparsity while maintaining performance comparable to that of fully connected networks, this paper presents CH-SNN, a four-stage dynamic sparse training framework for learning ultra-sparse spiking neural networks. The framework comprises: (1) sparse topology initialization, leveraging input correlation to initialize the network structure; (2) sparse weight initialization, which incorporates temporal activation sparsity, structural connection sparsity and neuronal threshold information of SNNs to initialize the weights of SNNs within a sparse network structure. (3) network pruning based on a removal score, combined with the removal of inactive neurons to improve information flow, and (4) network regrowth using the CH3-L3 score with a probabilistic strategy. These stages enable CH-SNN to achieve sparsification across all linear layers and enables effective training at ultra-high levels of sparsity conditions. Experimental results demonstrate that CH-SNN outperforms existing sparse SNNs training methods, achieving 97.75\% sparsity on MNIST with a 0.16\% accuracy improvement over the FC baseline. In addition, it realizes 98.84\% sparsity with a 2.27\% performance gain over the FC network while improving energy efficiency by approximately 97.5×. In summary, CH-SNN achieves performance comparable to FC networks even at ultra-high levels of sparsity, offering a promising solution for implementing edge AI on neuromorphic hardware.


\bibliography{iclr2026_conference}
\bibliographystyle{iclr2026_conference}

\appendix
\section{Appendix}
\subsection{Sparse spike weight initialization} \label{a.1}
For the sparse spiking neural network, we first introduce the sparse connection matrix $\displaystyle \mC \in \{0, 1\}^{m\times n}$, where 1 represents a connection exists and 0 represents no connection exists. For a sparse network, we have:
\begin{equation}
    \label{equ: comput_y}
    \displaystyle \vy^{(l)} = \mC^{(l)} \odot \mW^{(l)} \vx^{(l)} + \vb{(l)}
\end{equation}
where $\odot$ represents the Hadamard product, $\mW^{(l)} \in \sR^{m \times n}$ is the weight matrix, $\vx^{(l)} \in \sR^{n \times 1}$ denotes the input vector, $\vb^{(l)} \in \sR^{m \times 1}$ is the bias vector, $\vy^{(l)} \in \sR^{m \times 1}$ stands for the output vector, and $l$ represents the layer index. Next we assume that bias is 0 and the number of network layers is $L$, For the $i-\text{th}$ element $\evy^{(l)}_i$ of the output vector $\vy^{(l)}$, we discuss its variance:
\begin{equation}
\label{equ: var[y1]}
    Var[\evy^{(l)}_i] = Var[\sum^{n}_{j=1}\emC^{(l)}_{ij}\emW^{(l)}_{ij}\evx^{(l)}_j]
\end{equation}
$\mC$, $\mW$ and $\vx$ are independent of each other, All elements in the matrices $\mW$,$\mC$,$\vx$ are independently and identically distributed, and the three matrices have different distributions. Then the variance of $\evy^{(l)}_i$ can be expressed as follows:
\begin{equation}
\label{equ: var[y2]}
    Var[\evy^{(l)}_i] = n[(Var[\emC_{ij}] + \mu_{\emC}^2)(Var[\emW_{ij}] + \mu_{\emW}^2)(Var[\evx_{j}] + \mu_{\evx}^2)-\mu_{\emC}^2\mu_{\emW}^2\mu_{\evx}^2]
\end{equation}
where $\mu$ is the mean. For the elements of the sparse connection matrix $\mC$, we assume that it follows the Bernoulli distribution, then we have:
\begin{equation}
\label{equ: C distribution}
\emC_{ij}= \left\{
\begin{aligned}
1, \ p&=1-S_s \\
0, \ p&=S_s
\end{aligned}
\right. ,\quad  \mu_{\emC} = 1 - S_s, \quad Var[\emC_{ij}] = S_s(1-S_s)
\end{equation}
where $S_s$ is the structural connection sparsity. Based on previous work~\citep{kaiming}, we similarly define $\emW_{ij}$ to follow a zero-mean Gaussian distribution, which means $\mu_{\emW}=0$. The Equation~(\ref{equ: var[y2]}) can be changed to:
\begin{equation}
\label{equ: vary3}
    Var[\evy^{(l)}_i] = n(1-S_s)(Var[\emW_{ij}])(Var[\evx_{j}] + \mu_{\evx}^2)
\end{equation}
We expect to maintain the same variance of the input feature across layers, which means $Var[\evy^{(l)}_i] = Var[\evy^{(l-1)}_i] $. For the input layer ($l=1$), it can be expressed as $Var[\evy^{(1)}_i] = Var[\evx_i]$, because there is no activation function applied to the input feature. The input of the spiking neural network is 1 or 0, therefore, we can assume that $\evx_i$ follows the Bernoulli distribution with parameter $p=S_t$. We can get the variance of the $\emW_{ij}$ as follows:
\begin{equation}
    Var[\emW_{ij}^{(1)}] = \frac{S_t}{n(1-S_s)}
\end{equation}
where $S_t$ is the temporal activation sparsity, and $n$ is the dimension of the input.  For the rest of sparsely connected layers, $\evy^{(l)}_i$ follows a distribution with zero mean and symmetric about zero, since $\emW^{(l)}_{ij}$ follows a zero-mean Gaussian distribution and the input of every layer is zero or one, which will not affect the symmetry of $\emW^{(l)}_{ij}$. To simplify the analysis, we assume that $\evy^{(l)}_i$ also follows a zero-mean Gaussian distribution. In the spiking neural network, the activation function is not ReLU, but rather the step function, which can be expressed as:
\begin{equation}
\label{equ: step function}
\evx^{(l)}_i = U(\evy^{(l-1)}_{i}+V_{mem}-\theta)= \left\{
\begin{aligned}
1, \quad \evy^{(l-1)}_{i}+V_{mem} &\geq \theta \\
0, \quad \evy^{(l-1)}_{i}+V_{mem} &< \theta
\end{aligned}
\right.
\end{equation}
Where $\theta$ is the threshold of the spiking neural network. We approximate that $V_{mem}$ and $\evy^{(l-1)}_i$ follow the same distribution. Therefore, Equation~(\ref{equ: step function}) can be expressed as follows:
\begin{equation}
\label{equ:hard soft}
\evx^{(l)}_i = U(2\evy^{(l-1)}_{i}-\theta)= \left\{
\begin{aligned}
1, \quad \evy^{(l-1)}&\geq \theta/2 \\
0, \quad \evy^{(l-1)}&< \theta/2
\end{aligned}
\right.
\end{equation}
We expect to find a relationship between $Var[\evx^{(l)}_{j}] + \mu_{\evx}^2$ and $Var[\evy^{(l-1)}_j]$ that will simplify Equation~(\ref{equ: vary3}). Based on Equation~(\ref{equ:hard soft}), we have:
\begin{equation}
\label{equ: jifen}
    Var[\evx^{(l)}_{j}] + \mu_{\evx}^2 = \int_{\theta/2}^{+\infty} \frac{1}{\sqrt{2\pi Var[\evy^{(l-1)}_j]}} e^{-{x^2}/{2Var[\evy^{(l-1)}_j]}}dx= 1 - \Phi(\theta/2\sqrt{Var[\evy^{(l-1)}_j]})
\end{equation}
where $\Phi(\cdot)$ is the cumulative distribution function of the standard normal distribution. We expand Equation~(\ref{equ: jifen}) in a Taylor series around $\theta/2$ with $Var[\evy^{(l-1)}_j]$ as the variable and discard the higher-order terms:
\begin{equation}
\label{equ:Taylor}
    Var[\evx^{(l)}_{j}] + \mu_{\evx}^2 \approx 1 - \Phi(1) + \frac{2}{\sqrt{2\pi}\theta^2}e^{(-\frac{1}{2})}(Var[\evy^{(l-1)}_j] - \frac{\theta^2}{4}) \approx \frac{\sqrt{2}e^{-1/2}}{\sqrt{\pi}\theta^2}Var[\evy^{(l-1)}_j]
\end{equation}
Next, we substitute Equation~(\ref{equ:Taylor}) into Equation~(\ref{equ: vary3}), obtaining the following result:
\begin{equation}
\label{equ:var rest layers}
    Var[\emW_{ij}^{(l)}] = \frac{\theta^2\sqrt{\pi}}{\sqrt{2}e^{-1/2}n(1-S_s)}, (1< l <L)
\end{equation}
Finally, we assume that the output layer is fully connected, which means $S_s=0$, so we can obtain the following result:
\begin{equation}
\label{equ:var out layers}
    Var[\emW_{ij}^{(L)}] = \frac{\theta^2\sqrt{\pi}}{\sqrt{2}e^{-1/2}n}
\end{equation}
In summary, we can summarize the Sparse Spike Weight Magnitude Initialization (SSWI) as follows:
\begin{equation}
\label{equ:sswi_add}
    SSWI(\emW_{ij}^{(l)}) \sim \mathcal{N}(0,\sigma^2),\quad \sigma^2 = \left\{
\begin{aligned}
&\frac{S_t}{n(1-S_s)}, \quad (l=1) \\
&\frac{\theta^2\sqrt{\pi}}{\sqrt{2}e^{-1/2}n(1-S_s)},\quad ( 1<l<L) \\
& \frac{\theta^2\sqrt{\pi}}{\sqrt{2}e^{-1/2}n},\quad (l=L)
\end{aligned}
\right.
\end{equation}

\subsection{Explanation of CH3-L3} \label{a.2}
To illustrate the application of CH3-L3 for link prediction, consider a hypothetical network containing nodes $u$ and $v$ that are not directly connected. The CH3-L3 method evaluates potential links by analyzing all length-3 paths between $u$ and $v$ and incorporating the local community connectivity of intermediate nodes.

\begin{figure}[h]
    \centering
    \includegraphics[width=0.85\linewidth]{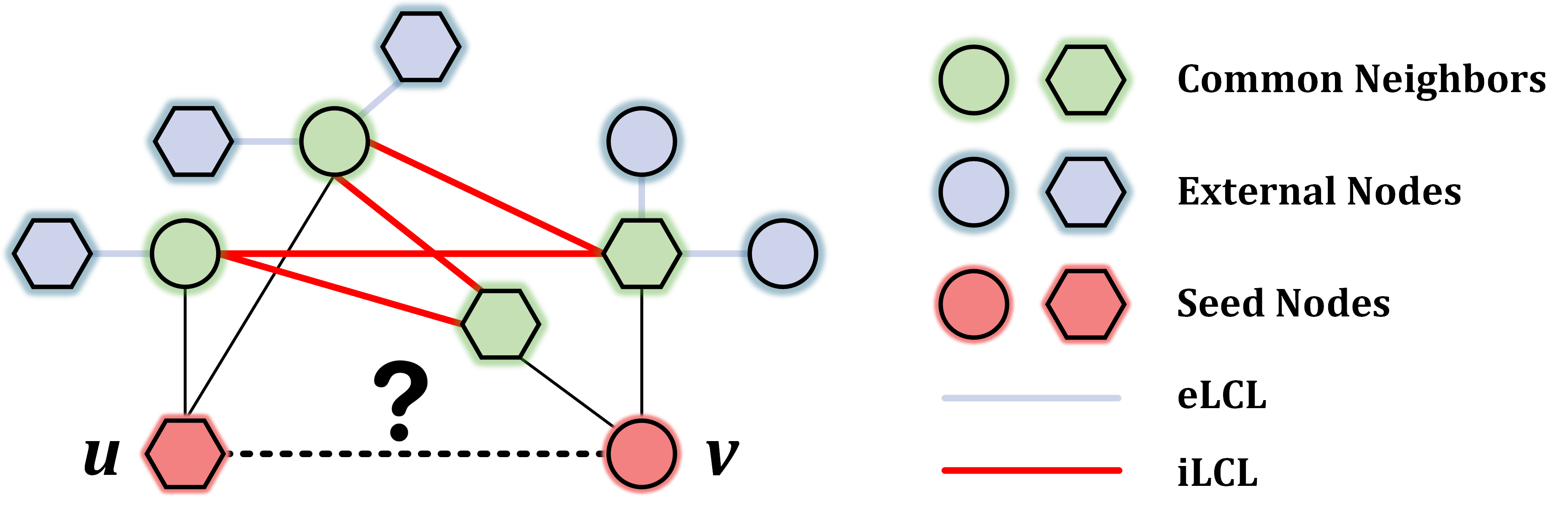}
    \caption{Example of link prediction using CH3-L3.}
    \label{fig:CH3-L3}
\end{figure}

We provide an illustrative example in Figure~\ref{fig:CH3-L3}. Red nodes represent seed nodes between which no direct connection is currently observed, but which have the potential to regenerate a link. Green nodes denote common neighbors—nodes directly connected to both seed nodes. Red edges indicate connections between common neighbors, referred to as internal Local Community Links (iLCL). Blue nodes represent external nodes outside the set of seed nodes and their common neighbors. Blue edges correspond to connections between common neighbors and these external nodes, termed external Local Community Links (eLCL). The calculation process of CH3-L3 is as follows:

\begin{equation}
\label{equ:a.ch3-l3}
    \textbf{CH3-L3}(u, v)= \sum_{z_1,z_2\in l3(u,v)}\frac{1}{\sqrt{(1+de_{z_{1}})\times (1+de_{z_{2}})}}
\end{equation}

where $u$ and $v$ denote two seed nodes, and $z_1$, $z_2$ represent two common neighbor nodes along a path of length 3 between $u$ and $v$. The terms $de_{z_1}$ and $de_{z_2}$ correspond to the number of eLCL associated with node $z_1$ and $z_2$, respectively. The CH3-L3 score is computed by summing the contributions from all length-3 paths between $u$ and $v$. Here, adding 1 to $de_{z_1}$ and $de_{z_2}$ prevents division by zero and ensures numerical stability when no external links are present.

\subsection{Experimental setup} \label{a.setup}
To evaluate the performance of CH-SNN, we conduct extensive experiments on multiple datasets, including MNIST~\citep{mnist}, CIFAR10~\citep{cifar10_100}, CIFAR100~\citep{cifar10_100}, N-MNIST~\citep{nmnist}, CIFAR10-DVS~\citep{dvsc10} and DVS-Gesture~\citep{dvsg}. Both the MNIST and N-MNIST datasets contain 10 classes of handwritten numbers (0–9), each consisting of 60,000 training samples and 10,000 test samples. DVS-Gesture comprises 11 types of gestures captured using an event-based camera. For these three datasets, we adopt the following network architecture: Input → \textbf{\color{red}{15}} (Channel Count)\textbf{\color{red}{Conv}} (Layer Type)\textbf{\color{red}{3$\times$3}}(Kernel Size) → AvgPool2$\times$2 → 40Conv3$\times$3 → AvgPool2$\times$2 → Flatten → 300 → Classes. For MNIST, a two-layer fully connected network is used: 784 → 1568 → 10. Since CIFAR10, CIFAR10-DVS and CIFAR100 share a similar data format, we employ the same network structure for both: Input → [128Conv3$\times$3]$\times$2 → MaxPool2$\times$2 → [256Conv3$\times$3]$\times$2 → MaxPool2$\times$2 → 512Conv3$\times$3 → Flatten → 512$\times$8$\times$8 → 512 → 10. All networks are configured with a time step of 8. Non-Spiking datasets (CIFAR10, CIFAR100, MNIST) are encoded using direct encoding. A uniform sparsity of 99\% is applied to all linear layers (except for the output layer). Weight updates are performed via surrogate gradient methods. Within each epoch, after completing weight training, CH-SNN performs pruning and regeneration according to the pruning ratio $\zeta$ followed by testing. The value of $\zeta$ decays cosine-annealingly to zero over the course of training, which can be expressed as follows:
\begin{equation}
\label{equ:a.zeta}
    \zeta = \frac{\zeta}{2} \times (1+\text{cos}(\frac{\pi\times epoch}{total\ epochs}))
\end{equation}
For the S-TP algorithm, we perform experiments on the CIFAR-10, MNIST, N-MNIST and DVS-Gesture datasets using a network structure defined as Input-2×Input-2×Input-Classes, with input sizes of 578 for N-MNIST, 784 for MNIST, 512 for CIFAR-10, and 2048 for DVS-Gesture. Throughout training, we set the learning rate to 0.0001, the dynamic pruning ratio $\zeta$ to 0.35, the batch size to 100, the target window size to 4, the number of epochs to 100, and the neuronal firing threshold to 4.

\subsection{Ablation experiment} \label{a.ablation}
We have conducted ablation studies to validate the effectiveness of our proposed SSCTI and SSWI. The results are summarized in Table~\ref{tab:ablation_study_stp}. When both SSCTI and SSWI are removed, the model fails to converge in most cases. Removing either SSCTI or SSWI individually leads to varying degrees of performance degradation. Notably, with a high level of sparsity of 99\%, the model becomes unstable and fails to train when SSWI is ablated. These ablation results demonstrate the critical importance and effectiveness of both SSCTI and SSWI in maintaining network performance under extreme sparsity. 
\begin{table}[h]
\centering
\caption{Ablation Study on SSCTI and SSWI (When SSCTI is ablated, we employ random structural initialization; when SSWI is removed, we use Kaiming initialization).}
\renewcommand{\arraystretch}{1}
\label{tab:ablation_study_stp}
\begin{tabular}{ccccccc}
\toprule[1.5pt]
\textbf{Dataset} & \textbf{SSCTI} & \textbf{SSWI} & \makecell{\textbf{99\%} \\ \textbf{Sparsity}} & \makecell{\textbf{95\%} \\ \textbf{Sparsity}} & \makecell{\textbf{80\%} \\ \textbf{Sparsity}} & \makecell{\textbf{70\%} \\ \textbf{Sparsity}} \\
\hline 
\multirow{4}{*}{\textbf{N-MNIST}}   
                    &               &              & \text{9.80\%}    & \text{9.80\%}   & \text{9.80\%}   & \text{91.99\%} \\
                    & \checkmark    &              & \text{9.80\%}    & \text{88.82\%}   & \text{96.95\%}   & \text{97.12\%} \\
                    &               & \checkmark   & \text{89.61\%}    & \text{94.74\%}   & \text{96.53\%}   & \text{96.87\%} \\
                    & \checkmark    & \checkmark   & \textbf{96.20\%}  & \textbf{97.21\%} & \textbf{97.29\%} & \textbf{97.22\%} \\
\hline
\multirow{4}{*}{\textbf{MNIST}}   
                    &               &              & \text{9.80\%}    & \text{9.80\%}   & \text{97.23\%}   & \text{97.78\%} \\
                    & \checkmark    &              & \text{70.41\%}    & \text{96.57\%}   & \text{97.57\%}   & \text{97.57\%} \\
                    &               & \checkmark   & \text{78.09\%}    & \text{96.81\%}   & \text{97.88\%}   & \text{97.91\%} \\
                    & \checkmark    & \checkmark   & \textbf{96.88\%}  & \textbf{97.56\%} & \textbf{98.11\%} & \textbf{98.00\%} \\
\hline
\multirow{4}{*}{\textbf{DVS-Gesture}}   
                    &               &              & \text{9.09\%}     & \text{9.09\%}    & \text{9.09\%}    & \text{9.09\%} \\
                    & \checkmark    &              & \text{9.09\%}     & \text{9.09\%}    & \text{78.79\%}   & \text{86.74\%} \\
                    &               & \checkmark   & \text{76.52\%}    & \text{87.12\%}   & \text{87.50\%}   & \text{86.74\%} \\
                    & \checkmark    & \checkmark   & \textbf{91.29\%}  & \textbf{91.29\%} & \textbf{88.64\%} & \textbf{89.02\%} \\
\hline
\multirow{4}{*}{\textbf{CIFAR-10}}   
                    &               &              & \text{10.00\%}     & \text{57.35\%}    & \text{78.89\%}    & \text{77.95\%} \\
                    & \checkmark    &              & \text{34.55\%}     & \text{76.83\%}    & \text{80.01\%}   & \text{78.19\%} \\
                    &               & \checkmark   & \text{77.40\%}    & \text{81.62\%}   & \text{80.62\%}   & \text{80.65\%} \\
                    & \checkmark    & \checkmark   & \textbf{78.94\%}  & \textbf{82.84\%} & \textbf{81.73\%} & \textbf{81.42\%} \\
\bottomrule[1.5pt]     
\end{tabular}
\end{table}

\subsection{Sensitivity test} \label{a.sensitivity}
We have conducted a sensitivity analysis of the hyperparameters in CH-SNN, focusing primarily on the learning rate, batch size, dynamic pruning ratio, and static sparsity rate, to evaluate the model's performance under variations in these parameters.

\begin{table}[h]
\centering
\caption{Learning rate sensitivity experiment.}
\renewcommand{\arraystretch}{1.1}
\label{tab:lr}
\begin{tabular}{cccccc}
\toprule[1.5pt]
\textbf{Dataset} & \textbf{lr-0.01}  & \textbf{lr-0.005} & \textbf{lr-0.001} & \textbf{lr-0.0005} & \textbf{lr-0.0001} \\
\hline  
N-MNIST         & 83.08\%      & 85.59\%            & 95.84\%           & 96.87\%           & 96.91\%       \\
DVS-Gesture     & 65.91\%      & 75.76\%            & 89.77\%           & 89.77\%           & 89.77\%       \\
MNIST           & 67.18\%      & 85.83\%            & 94.98\%           & 96.19\%           & 97.98\%       \\
CIFAR-10        & 78.74\%      & 79.96\%            & 82.84\%           & 82.73\%           & 82.54\%       \\
\hline
N-MNIST(FC)         & 69.65\%      & 80.90\%            & 88.16\%           & 96.31\%           & 96.38\%       \\
DVS-Gesture(FC)     & 68.56\%      & 68.18\%            & 85.98\%           & 88.64\%           & 89.02\%       \\
MNIST(FC)           & 79.90\%      & 84.58\%            & 92.43\%           & 94.12\%           & 97.29\%       \\
CIFAR-10(FC)        & 59.86\%      & 55.49\%            & 80.14\%           & 81.92\%           & 80.67\%       \\
\bottomrule[1.5pt]     
\end{tabular}
\end{table}

\textbf{Learning Rate (LR).} 
We train CH-SNN using different learning rates (0.01, 0.005, 0.001, 0.0005, 0.0001) and record its performance, as summarized in Table~\ref{tab:lr}. The results indicate that as the learning rate increases, the model exhibits a noticeable decline in performance. Through further analysis, we conclude that this performance degradation stems from the S-TP algorithm: during weight updates, an excessively large learning rate leads to oversized training steps, preventing convergence to an optimal solution. This is validated by conducting a learning rate sensitivity experiment on a fully connected network, where a similar performance drop is observed, as shown in Table~\ref{tab:lr}.



\textbf{Batch Size (BS).}
We employ different batch sizes to train CH-SNN, and the experimental results presented in Table~\ref{tab:bs} show that variations in batch size did not cause significant changes in its performance.

\begin{table}[h]
\centering
\caption{Batch size sensitivity experiment.}
\renewcommand{\arraystretch}{1.1}
\label{tab:bs}
\begin{tabular}{cccccc}
\toprule[1.5pt]
\textbf{Dataset} & \textbf{BS-16}  & \textbf{BS-32} & \textbf{BS-50} & \textbf{BS-64} & \textbf{BS-100} \\
\hline
N-MNIST         & 96.88\%      & 96.86\%            & 96.85\%           & 97.00\%           & 96.91\%       \\
DVS-Gesture     & 89.77\%      & 89.39\%            & 89.02\%           & 89.39\%           & 89.77\%       \\
MNIST           & 98.08\%      & 98.09\%            & 98.11\%           & 98.07\%           & 97.98\%       \\
CIFAR-10        & 82.64\%      & 81.95\%         & 81.88\%           & 81.74\%           & 82.84\%       \\
\bottomrule[1.5pt]   
\end{tabular}
\end{table}
\textbf{Dynamic Pruning Ratio ($\zeta$).}
To evaluate the impact of different pruning rate strategies on model performance, we first test the dynamic pruning rate strategy as defined in Equation (\ref{equ:a.zeta}). We adjust various decay starting points (0.5, 0.4, 0.3, 0.2, 0.1), and the experimental results are shown in Table~\ref{tab:dynamic zeta}. It can be observed that CH-SNN exhibits negligible performance variation across different initial pruning rates, demonstrating strong stability in response to changes in the pruning rate. The model consistently achieves strong performance under the predefined sparsity targets.

\begin{table}[h]
\centering
\caption{Dynamic removal rate sensitivity experiment.}
\renewcommand{\arraystretch}{1.1}
\label{tab:dynamic zeta}
\begin{tabular}{cccccc}
\toprule[1.5pt]
\textbf{Dataset} & \textbf{$\zeta$-0.5}  & \textbf{$\zeta$-0.4} & \textbf{$\zeta$-0.3} & \textbf{$\zeta$-0.2} & \textbf{$\zeta$-0.1} \\
\hline  
N-MNIST         & 97.14\%      & 96.91\%            & 96.96\%           & 97.04\%           & 97.10\%       \\
DVS-Gesture     & 89.39\%      & 90.53\%            & 89.02\%           & 88.64\%           & 89.39\%       \\
MNIST           & 97.98\%      & 97.98\%            & 97.98\%           & 97.98\%           & 97.98\%       \\
CIFAR-10        & 82.64\%      & 82.24\%            & 82.84\%           & 83.03\%           & 82.75\%             \\
\bottomrule[1.5pt]        
\end{tabular}
\end{table}
\textbf{Static Pruning Ratio ($\zeta$).}
Similarly, we evaluate a static pruning rate strategy. Unlike the dynamic approach, the static pruning rate remains constant at its initial value throughout training. We test multiple starting values for the static pruning rate, and the results are presented in Table~\ref{tab:static zeta}. CH-SNN shows minimal performance variation across different static pruning rates. It is worth noting that compared to the dynamic pruning strategy, the static approach generally leads to a slight decrease in overall accuracy.

\begin{table}[h]
\centering
\caption{Static removal rate sensitivity experiment.}
\renewcommand{\arraystretch}{1.1}
\label{tab:static zeta}
\begin{tabular}{cccccc}
\toprule[1.5pt]
\textbf{Dataset} & \textbf{$\zeta$-0.5}  & \textbf{$\zeta$-0.4} & \textbf{$\zeta$-0.3} & \textbf{$\zeta$-0.2} & \textbf{$\zeta$-0.1} \\
\hline  
N-MNIST         & 96.90\%      & 96.87\%            & 96.84\%           & 96.81\%           & 97.10\%       \\
DVS-Gesture     & 88.64\%      & 89.39\%            & 89.02\%           & 89.02\%           & 88.64\%       \\
MNIST           & 97.98\%      & 97.98\%            & 97.98\%           & 97.98\%           & 97.98\%        \\
CIFAR-10        & 82.24\%      & 82.26\%            & 82.53\%           & 82.63\%           & 82.25\%       \\
\bottomrule[1.5pt]        
\end{tabular}
\end{table}

\subsection{Node sparsity} \label{a.node_sparsity}
In the CH-SNN framework, neurons that are unilaterally or bilaterally disconnected (i.e., without any incoming or outgoing links) are regarded as inactive neurons. Since these inactive neurons lose the ability to transmit information, they may hinder information flow throughout the network. We assume that such inactive neurons are unable to regrow new links during the network regrowth stage. Therefore, during the chain removal step, we permanently remove them from the network. As illustrated in Figure~\ref{fig:CSTI} Stage 3, this process enhances node sparsity. We compare CH-SNN with SD-SNN, an existing open-source method, as shown in Table~\ref{tab:node_sparsity}.

\begin{table}[h] 
\caption{Node sparsity of different methods on  CIFAR-10, CIFAR-100, MNIST, N-MNIST and DVS-Gesture  datasets.}
\centering
\renewcommand{\arraystretch}{1.05}
\label{tab:node_sparsity}
\begin{tabular}{lllccc}
\toprule[1.5pt]
\textbf{Dataset} & \textbf{Method} & \textbf{Network} & \textbf{Node sparsity} \\ 
\midrule[0.8pt] 
\multirow{2}{*}{\textbf{CIFAR-10}} 
                    & SD-SNN            & 6Conv 2FC                 & 0.28\%             \\
                    & \textbf{CH-SNN}            & \textbf{6Conv 2FC}                 & \textbf{1.21\%}               \\
\hline

\multirow{2}{*}{\textbf{CIFAR-100}} 
                   & SD-SNN            & 6Conv 2FC                  & 0.46\%               \\
                   & \textbf{CH-SNN}            & \textbf{6Conv 2FC}                  & \textbf{0.69\%}              \\
\hline
\multirow{4}{*}{\textbf{MNIST}} 
                    & SD-SNN            & 2Conv 2FC                 & 0.02\%                \\
                    & \textbf{CH-SNN}            & \textbf{2Conv 2FC}                 & \textbf{0.04\%}               \\
                    & SD-SNN            & 2FC                       & 0.77\%                \\
                    & \textbf{CH-SNN}            & \textbf{2FC}                       & \textbf{2.03\%}               \\
\hline

\multirow{2}{*}{\textbf{N-MNIST}} 
                    & SD-SNN            & 2Conv 2FC             & 6.32\%                \\
                    & \textbf{CH-SNN}            & \textbf{2Conv 2FC}             & \textbf{8.09\%}                \\
\hline
\multirow{2}{*}{\textbf{DVS-Gesture}} 
                    & SD-SNN            & 2Conv 2FC             & 1.28\%                \\
                    & \textbf{CH-SNN}            & \textbf{2Conv 2FC}             & \textbf{6.35\%}              \\
\bottomrule[1.5pt] 
\end{tabular}
\end{table}

\subsection{Limitation and future challenges} \label{a.limitation}
The primary limitation of CH-SNN is its current restriction to sparsifying only linear layers within spiking neural networks, while convolutional layers remain unaffected. The reason CH-SNN achieves significant sparsity increases and maintain performance comparable to that of fully connected networks lies in the fact that linear layers typically constitute the majority of synaptic connections in most architectures—whether in spiking convolutional networks or newer transformer-based models such as Spikformer. By sparsifying these dominant linear layers, CH-SNN elevates the overall network sparsity to a high level. A key direction for future work is therefore to extend the method to support ultra-sparse training in convolutional layers.

Another limitation pertains to the computational overhead associated with the CH3-L3 scoring process. While CH-SNN effectively reduces energy consumption on neuromorphic processors in settings that do not require continual learning, its application in on-device continual learning scenarios may introduce additional computational and memory burdens. In such cases, the integration of dedicated hardware for efficient CH3-L3 calculation would be necessary to mitigate these overheads.

\subsection{Usage of large language models.} \label{a.llm}
 In the process of preparing this manuscript, we utilized the DeepSeek large language model to assist in polishing the English writing and refining the wording of the Abstract, Introduction and Conclusion sections. The core ideas, theoretical contributions, experimental design, data analysis, and results remain entirely the work of the authors. The authors take full responsibility for the entire content of this paper.

\end{document}

%% file: math_commands.tex
\usepackage{amsmath,amsfonts,bm}

\def\eqref#1{equation~\ref{#1}}

\def\1{\bm{1}}

\def\vb{{\bm{b}}}

\def\vx{{\bm{x}}}
\def\vy{{\bm{y}}}

\def\evx{{x}}
\def\evy{{y}}

\def\mC{{\bm{C}}}

\def\mW{{\bm{W}}}

\DeclareMathAlphabet{\mathsfit}{\encodingdefault}{\sfdefault}{m}{sl}
\SetMathAlphabet{\mathsfit}{bold}{\encodingdefault}{\sfdefault}{bx}{n}

\def\sR{{\mathbb{R}}}

\def\emC{{C}}

\def\emW{{W}}

